\pdfoutput=1

\documentclass[11pt]{article}

\usepackage[final]{acl}

\usepackage{times}
\usepackage{latexsym}

\usepackage[T1]{fontenc}

\usepackage[utf8]{inputenc}

\usepackage{microtype}

\usepackage{inconsolata}

\usepackage{url}
\usepackage{amsmath}
\usepackage{booktabs}
\usepackage{multirow}
\usepackage{makecell}
\usepackage{graphicx}
\usepackage{tikz}
\usepackage{pgfplots}
\usepackage{enumitem}
\usepackage{amssymb}
\usepackage{pifont}
\usepackage{xcolor}
\usepackage{colortbl}
\usepackage{soul}
\usepackage{siunitx}

\usetikzlibrary{shadows}
\usetikzlibrary{plotmarks}
\usetikzlibrary{patterns}
\usetikzlibrary{pgfplots.groupplots}
\usetikzlibrary{arrows}
\usetikzlibrary{decorations}
\usetikzlibrary{decorations.pathreplacing}
\usetikzlibrary{backgrounds}
\usetikzlibrary{fit}
\usetikzlibrary{positioning}
\usetikzlibrary{calc}
\usetikzlibrary{shapes.multipart}

\definecolor{ugreen}{cmyk}{1,0,1,0.498}
\definecolor{lyyblue}{cmyk}{0.8278,0.3333,0,0.2941}
\definecolor{lyygreen}{cmyk}{0.6813,0,0.725,0.3725}
\definecolor{lyyred}{cmyk}{0,0.8855,0.8767,0.1098}
\definecolor{dblue}{cmyk}{1,0.5487,0,0.5569}

\definecolor{myred}{HTML}{E33222}

\newcommand{\cmark}{\ding{52}}%
\newcommand{\xmark}{\ding{56}}%

\definecolor{gr}{RGB}{0, 146, 0}
\newcommand{\gr}[1]{\textcolor{gr}{#1}}
\newcommand{\red}[1]{\textcolor{myred}{#1}}

\newcommand{\org}[1]{\textcolor{orange}{#1}}
\newcommand{\blue}[1]{\textcolor{blue}{#1}}

\newcommand{\gray}[1]{\textcolor{lightgray}{#1}}

\newcommand{\method}{AttrLoRA}
\newcommand{\dataset}{MuSiQue-Attribute}

\title{Making Long-Context Language Models Better Multi-Hop Reasoners}

\author{Yanyang Li$^{1}$, Shuo Liang$^{1,2}$, Michael R. Lyu$^1$, Liwei Wang$^1$\thanks{Corresponding author.}\\
$^1$Department of Computer Science and Engineering, The Chinese University of Hong Kong \\
$^2$Shanghai AI Laboratory\\
\texttt{\{yyli21,sliang23,lyu,lwwang\}@cse.cuhk.edu.hk}
}

\begin{document}
\maketitle
\begin{abstract}
Recent advancements in long-context modeling have enhanced language models (LMs) for complex tasks across multiple NLP applications. Despite this progress, we find that these models struggle with multi-hop reasoning and exhibit decreased performance in the presence of noisy contexts. In this paper, we introduce \textit{Reasoning with Attributions}, a novel approach that prompts LMs to supply attributions for each assertion during their reasoning. We validate our approach through experiments on three multi-hop datasets, employing both proprietary and open-source models, and demonstrate its efficacy and resilience. Furthermore, we explore methods to augment reasoning capabilities via fine-tuning and offer an attribution-annotated dataset and a specialized training strategy.
Our fine-tuned model achieves competitive performance on multi-hop reasoning benchmarks, closely paralleling proprietary LMs such as ChatGPT and Claude-instant\footnote{The dataset, model, and code are publicly available at \url{https://github.com/LaVi-Lab/LongContextReasoner}.}.
\end{abstract}

\section{Introduction}
\label{sec:intro}



The field of long-context modeling has garnered significant attention due to its importance in applications that demand extensive comprehension and generation capabilities~\cite{DBLP:conf/nips/LewisPPPKGKLYR020,DBLP:journals/corr/abs-2306-03091}. Techniques for long-context modeling~\cite{DBLP:journals/corr/abs-2306-15595,DBLP:journals/corr/abs-2309-00071,chen2023longlora} have been proposed with encouraging results on established benchmarks~\cite{DBLP:journals/corr/abs-2307-11088,DBLP:journals/corr/abs-2308-14508}.

Nevertheless, we have identified a gap in the performance of these models when it comes to multi-hop reasoning tasks, where a model must navigate and synthesize information from disparate sources to answer complex questions. Evidence from key benchmarks such as LongBench~\cite{DBLP:journals/corr/abs-2308-14508}, as well as our experimental results in Section~\ref{sec:result}, indicate that these long-context LMs underperform compared to leading multi-hop reasoning systems~\cite{zhang2023beam}. The reasons for this shortfall in multi-hop reasoning effectiveness are not yet fully understood.

We contend that the limitations in multi-hop reasoning observed in long-context LMs stem from two main issues: The inability to discern pertinent information within noisy contexts~\cite{liu2023lost} and the struggle to incorporate knowledge within the context effectively, particularly for smaller-scale models~\cite{DBLP:conf/emnlp/ZhengLDFWXC23}. To address these challenges, we introduce \textit{Reasoning with Attributions}, a methodology that compels LMs to substantiate their reasoning by linking assertions to relevant context segments, such as citations~\cite{gao-etal-2023-enabling} or direct quotations~\cite{DBLP:journals/corr/abs-2203-11147}. This approach not only guides LMs to perform targeted information retrieval to identify the position of relevant contexts, thereby reducing noise, but also ensures their responses are well-grounded in the source material. Our preliminary and comprehensive experimental findings, detailed in Sections \ref{sec:pilot} and \ref{sec:result}, confirm the efficacy and resilience of this method across various multi-hop reasoning benchmarks.

Despite these advancements, smaller long-context LMs exhibit continued difficulties in reasoning. We explore the potential for these models to improve through learning to reason and attribute simultaneously. Utilizing ChatGPT~\cite{DBLP:conf/nips/BrownMRSKDNSSAA20} to annotate the multi-hop reasoning dataset MuSiQue~\cite{trivedi-etal-2022-musique}, we create a specialized dataset \textit{\dataset{}} for fine-tuning models in this dual capacity. We propose a potent learning strategy that leverages multi-task learning and data augmentation to fully exploit these annotations. Our experiments with five long-context LMs across three multi-hop reasoning datasets and two general instruction-following datasets reveal that our fine-tuned Vicuna-7B model~\cite{DBLP:journals/corr/abs-2306-05685} surpasses similar-scale baselines by a substantial margin, i.e., more than 20 points on average, and even outperforms ChatGPT and Claude-instant on MuSiQue, albeit with a slight trade-off in other capabilities. This study illuminates a promising avenue to enhance the multi-hop reasoning capabilities of long-context LMs through a focus on attributions.


Our contributions can be summarized as follows:
\begin{itemize}[noitemsep, nolistsep]
\item We introduce \textit{Reasoning with Attributions}, an innovative reasoning paradigm that enhances both the performance and robustness of long-context LMs in multi-hop reasoning tasks.
\item We provide \textit{\dataset{}}, an attribution-annotated multi-hop reasoning dataset to support further research in this domain.
\item We develop a bespoke learning strategy that incorporates novel auxiliary tasks for multi-task learning and employs tailored data augmentation techniques.
\item Through rigorous testing on three multi-hop reasoning datasets and two general instruction-following benchmarks with both proprietary and open-source models, we demonstrate that our fine-tuned Vicuna-7B model achieves comparable multi-hop reasoning performance to ChatGPT with minimal impact on other capabilities. Additionally, our model shows resilience against varying degrees of contextual noise, underscoring the effectiveness of our methods.
\end{itemize}

\section{Reasoning with Attributions}

\begin{table}[t!]
    \centering
    {\small
    \begin{tabular}{p{0.45\textwidth}}
        \toprule
        \textbf{Instruction}: Write an accurate and concise answer for \ldots\\
        {\color{brown}<Retrieve for the question>}\\
        Document [1](Title: David Myles (musician)): \ldots\\
        Document [2](Title: Jamal Plays Jamal): \ldots\\
        Document [3](Title: Top and Bottom Brass): \ldots\\
        \gray{(Other retrieved documents are omitted.)}\\
        \\
        \textbf{Question}: What is the genre of the record label of the band that performed on the Crush Tour?\\
        \\
        \textbf{Answer}:\\
        \midrule
        \textit{CoT}:\\
        The Crush Tour is performed by the band Bon Jovi. The record label of Bon Jovi is Island Records. The genre of Island Records is jazz. The answer is: \gr{jazz \cmark}\\
        \midrule
        \textit{CoC}:\\
        The Crush Tour is performed by the band Bon Jovi \org{[8]}. The record label of Bon Jovi is Island Records \org{[17]}. The genre of Island Records is jazz \org{[19]}. The answer is: \gr{jazz \cmark}\\
        \midrule
        \textit{CoQ}:\\
        The Crush Tour is performed by the band Bon Jovi (\blue{``The Crush Tour is a third concert''} \org{[8]}). The record label of Bon Jovi is Island Records (\blue{``Bounce is the eighth studio album by American''} \org{[17]}). The genre of Island Records is jazz (\blue{``The Antidote is the debut album by English jazz''} \org{[19]}). The answer is: \gr{jazz \cmark}\\
        \bottomrule
    \end{tabular}
    }
    \caption{An example of CoT and two of our reasoning with attribution methods: CoC and CoQ. We highlight the differences between these methods, e.g., answers are marked in \gr{green}, citations are marked in \org{orange} and quotes are marked in \blue{blue}.}
    \label{tab:example}
\end{table}

\subsection{Pilot Study}
\label{sec:pilot}


The challenge of large language models becoming mired in irrelevant contexts, known as the ``Lost in the Middle'' phenomenon, has been documented across various NLP tasks, such as multi-document QA~\cite{liu2023lost} and mathematical reasoning~\cite{DBLP:conf/icml/ShiCMSDCSZ23}. This issue is also apparent in multi-hop reasoning, which we illustrate later in Figure \ref{fig:noise}. Prior research has noted this problem but has not decoded the underlying mechanisms. For example, while \citet{liu2023lost} found that introducing the query before the context can aid in better information retrieval from the context, they did not achieve an improvement in QA performance using this query-aware approach. As we suggest in Section~\ref{sec:intro}, the reasons might extend beyond mere retrieval challenges to include complications in effectively applying the retrieved knowledge.


To tackle the issues outlined earlier, we introduce \textit{Reasoning with Attributions}, a strategy that mandates language models to link the claims made during reasoning to specific sections of the provided context. This implicit requirement effectively decomposes a complex multi-hop question into two more manageable tasks: Pinpointing pertinent information within the context and constructing well-founded claims based on that information.

We adapt the concept of Chain-of-Thought (CoT)~\cite{DBLP:conf/nips/Wei0SBIXCLZ22} reasoning to create two distinct variants aligned with our attribution-based approach: \textbf{Chain-of-Citation} (CoC) and \textbf{Chain-of-Quote} (CoQ). In CoC, models are prompted to reference citations corresponding to each step of the reasoning chain. CoQ goes further by requiring models to include direct quotations from the cited material for each reasoning step. An illustrative example highlighting the nuances between these methods is provided in Table~\ref{tab:example}.


The results of our preliminary study (Please refer to Section~\ref{sec:exp} for the setup), detailed in Table~\ref{tab:pilot}, compare the efficacy of CoT, CoC, and CoQ when applied to two proprietary long-context LMs: ChatGPT~\cite{DBLP:conf/nips/BrownMRSKDNSSAA20} and Claude-instant~\cite{DBLP:journals/corr/abs-2204-05862}. Without further notice, ChatGPT always refers to \texttt{gpt-3.5-turbo-1106} and \texttt{claude-instant-1.2} for Claude-instant in this work. The findings suggest that both CoC and CoQ generally yield improvements over CoT, indicating that attribution-based reasoning enhances the precision and coherence of the models' reasoning processes. CoQ appears to slightly underperform CoC, likely due to the increased complexity of producing exact quotations.

It is noteworthy that even in instances where CoT reduces the Answer Only (AO) performance, CoC is able to not only mitigate this decline but also surpass the AO baseline. This demonstrates the potential of CoC as a robust reasoning method. The success of our approach with various open-sourced models is further elaborated upon in Section~\ref{sec:result}. Based on these insights, we adopt CoC as our primary reasoning format in subsequent sections.

\begin{table}[t!]
    \centering
    \setlength{\tabcolsep}{4pt}
    {\small
    \begin{tabular}{l|cc|cc|cc}
        \toprule
        \makecell[c]{\multirow{3}*{\textbf{Model}}} & \multicolumn{2}{c|}{\textbf{MuSiQue}} & \multicolumn{2}{c|}{\textbf{2Wiki}} & \multicolumn{2}{c}{\textbf{HotpotQA}} \\
        \cmidrule{2-7}
        & \textbf{EM} & \textbf{F1} & \textbf{EM} & \textbf{F1} & \textbf{EM} & \textbf{F1} \\
        \midrule
        \multicolumn{7}{c}{\cellcolor{lightgray}\textit{ChatGPT} (\texttt{gpt-3.5-turbo-1106})} \\
        \midrule
        + AO & 15.8 & 26.9 & 46.2 & 57.2 & 51.0 & 65.4 \\
        + CoT & 36.2 & 50.1 & 55.2 & 70.1 & 56.8 & 71.2 \\
        + CoC & \textbf{37.0} & 51.0 & \textbf{55.4} & \textbf{71.1} & \textbf{58.6} & \textbf{73.4} \\
        + CoQ & 36.4 & \textbf{51.3} & 54.0 & 68.7 & 55.4 & 70.2 \\
        \midrule
        \multicolumn{7}{c}{\cellcolor{lightgray}\textit{Claude-instant} (\texttt{claude-instant-1.2})} \\
        \midrule
        + AO & 26.2 & 39.4 & 47.0 & 57.5 & \textbf{54.4} & \textbf{68.4} \\
        + CoT & 26.0 & 37.9 & 40.8 & 52.3 & 20.2 & 26.3 \\
        + CoC & \textbf{32.2} & \textbf{46.2} & \textbf{53.4} & \textbf{67.0} & 54.2 & 68.3 \\
        + CoQ & 30.2 & 45.9 & 49.8 & 62.1 & 50.8 & 65.0 \\
        \bottomrule
    \end{tabular}
    }
    \caption{Exact-Match (EM) and F1 scores of ChatGPT and Claude-instant with 5-shot prompting on multi-hop reasoning datasets, e.g., MuSiQue, 2WikiMultiHopQA (2Wiki for short) and HotpotQA. The best results are in \textbf{bold}. AO means models predict answers only.}
    \label{tab:pilot}
\end{table}

\subsection{Dataset Curation}



Our analysis, evidenced by the data in Tables \ref{tab:pilot} and \ref{tab:main}, confirms that while reasoning with attributions holds promise, smaller open-source long-context language models significantly underperform compared to their proprietary counterparts in multi-hop reasoning tasks. To address this, we investigate whether training these models to perform attributions can boost their reasoning capabilities.

A hurdle in this process is the lack of attribution annotations within existing multi-hop reasoning benchmarks. To bridge this gap, we have generated new annotations by prompting ChatGPT with 5-shot CoQ. This has been done to create CoT with attributions for 5,000 instances randomly selected from the answerable training set of the MuSiQue dataset~\cite{trivedi-etal-2022-musique}. Although CoC generally outperforms CoQ, we chose CoQ for annotation because it provides more detailed information. This richness is beneficial not only for evaluating the quality of the annotations but also proves advantageous for the fine-tuning processes discussed in Section~\ref{sec:learning}.

\begin{table}[t!]
    \centering
    {\small
    \begin{tabular}{l|r}
        \toprule
        \makecell[c]{\textbf{Error Type}} & \makecell[c]{\textbf{Portion}} \\
        \midrule
        Incorrect Answer & 58.44\% \\ 
        Non-Existent Attributions & 12.56\% \\ 
        Incorrect Citations & 9.80\% \\ 
        Repeated Citations  & 6.35\% \\ 
        Extreme Quotes  & 10.55\% \\ 
        \bottomrule
    \end{tabular}
    }
    \caption{Incidence rates of different error types.}
    \label{tab:filter}
\end{table}

\begin{table}[t!]
    \centering
    {\small
    \begin{tabular}{l|r}
        \toprule
        \makecell[c]{\textbf{Entry}} & \makecell[c]{\textbf{Value}} \\
        \midrule
        \#Max Words per Sample & 3385 \\
        \#Mean Words per Sample & 1809.10 \\
        \#Averaged Words per CoT Step  & 11.64 \\
        \#Averaged Words per Quote  & 16.60 \\
        \#Totoal Samples & 1358 \\
        \quad 2-Hop Samples [\%] & 82.18\% \\
        \quad 3-Hop Samples [\%] & 14.06\% \\
        \quad 4-Hop Samples [\%] & 3.76\% \\
        \bottomrule
    \end{tabular}
    }
    \caption{Statistics of \dataset{}.}
    \label{tab:train}
\end{table}


After generating the annotations, we implemented a filtering process to exclude annotations with any of the following errors (Please refer to Appendix~\ref{sec:implement} for implementation details):
\begin{itemize}[noitemsep, nolistsep]
\item \textbf{Incorrect Answer}: The model's predicted answer does not align with the reference answer, which typically indicates an erroneous CoT.
\item \textbf{Non-Existent Attributions}: Fabricated citations or quotes that do not correspond to the actual context are indicative of model hallucination.
\item \textbf{Incorrect Citations}: Citations do not match the manually identified supporting facts, suggesting flawed attributions.
\item \textbf{Repeated Citations}: Redundant citations contravene the multi-hop requirement of sourcing from multiple documents.
\item \textbf{Extreme Quotes}: Quotes that are either too terse (under five words) or excessively lengthy (spanning an entire document) lack utility.
\end{itemize}


Table \ref{tab:filter} presents the substantial incidence rates of each error type, which could negatively impact fine-tuning effectiveness. After filtering, we obtain a training dataset of 1,358 samples, referred to as \textit{\dataset{}}. The statistics of the \dataset{} training set are outlined in Table \ref{tab:train}. It is important to note that the hop distribution in \dataset{} is skewed. This skewness arises both because generated CoT for questions with more hops is more prone to errors and because such questions represent a smaller fraction of the original MuSiQue training set. Additionally, we conducted a human evaluation of the \dataset{} quality in Appendix \ref{sec:human}.

\section{Learning to Attribute in Reasoning}
\label{sec:learning}


One intuitive approach to enhancing the multi-hop reasoning capabilities of LMs is to fine-tune them on our curated \dataset{}, thereby teaching them to integrate attribution into their reasoning processes, specifically to generate CoC. Despite the simplicity of this method, our subsequent analysis in Section~\ref{sec:analysis} demonstrates that this direct approach fails to produce robust results.

\paragraph{Multi-Task Learning.}

\begin{figure}[t!]
\begin{center}
\includegraphics[width=\linewidth]{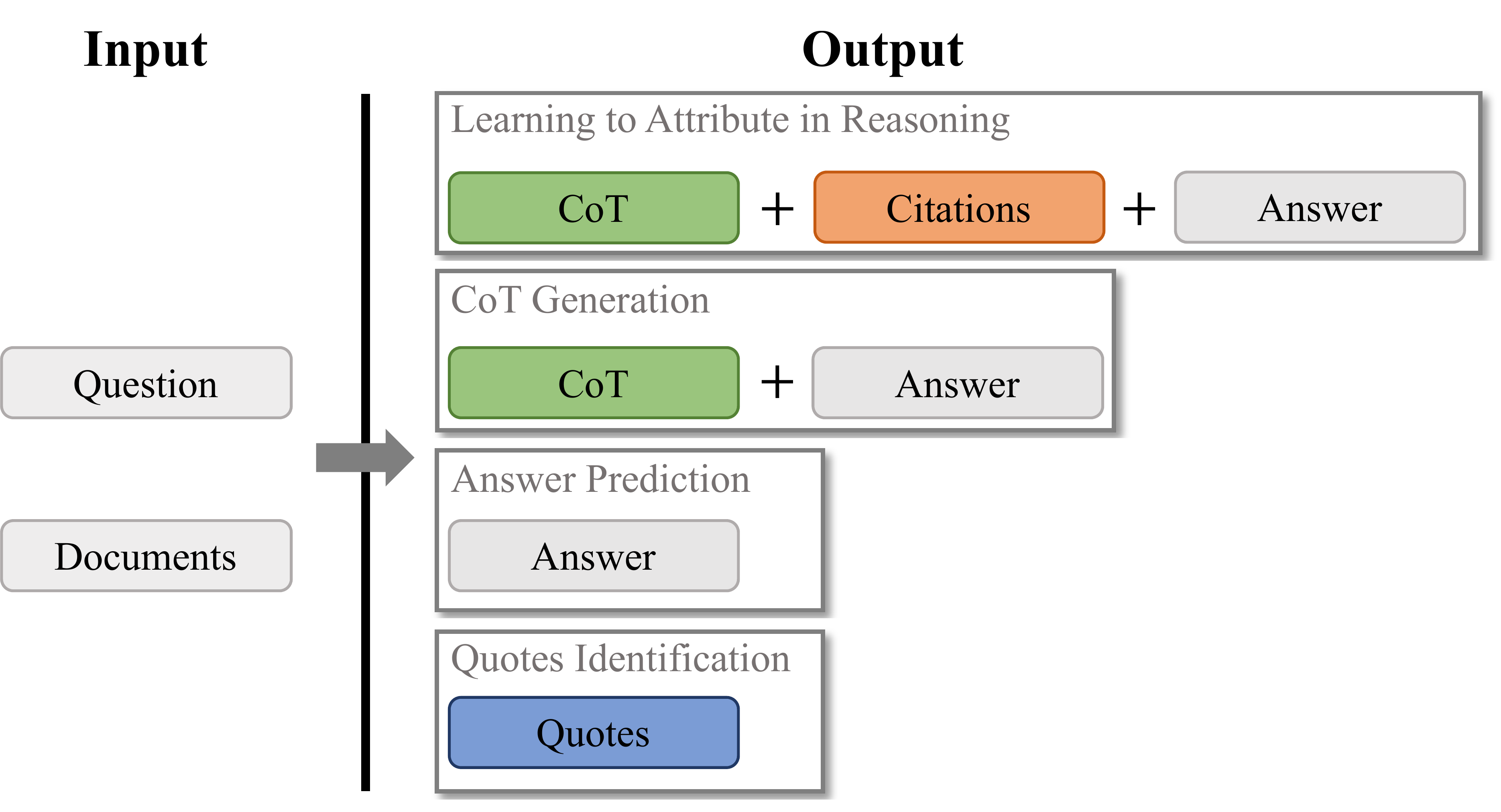}
\end{center}
\caption{Comparison of the proposed auxiliary tasks.} 
\label{fig:tasks}
\end{figure}


Beyond simply fine-tuning LMs on the \dataset{} to learn to attribute in reasoning (denoted as \textbf{LA}), we propose three auxiliary tasks that serve as simplified analogs of LA. These tasks are designed to train LMs in conjunction with LA to enhance their proficiency in attribution-based reasoning:
\begin{itemize}[noitemsep, nolistsep]
\item \textbf{Answer Prediction} (\textbf{AP} for short): This task focuses on direct answer prediction without the need for an explicit reasoning process. AP is intended to help LMs internalize the reasoning needed for straightforward questions where CoT is not required.
\item \textbf{CoT Generation} (\textbf{CG} for short): In the CG task, models are trained to generate a CoT before providing an answer. This is aimed at developing LMs' abilities to reason explicitly and methodically across multiple pieces of information for complex questions.
\item \textbf{Quotes Identification} (\textbf{QI} for short): This task trains models to pinpoint critical quotes for reasoning. QI is designed to fine-tune the ability of LMs to filter out irrelevant details and zero in on the pertinent segments of text, thereby sharpening the accuracy of reasoning.
\end{itemize}
Figure \ref{fig:tasks} illustrates the distinctions between our primary LA task and the three auxiliary tasks.

\paragraph{Data Augmentation.}


A recognized limitation of direct fine-tuning on our \dataset{} is the potential for models to develop biases, such as favoring certain locations of relevant documents~\cite{liu2023lost}, sensitive to a fixed number of documents, or accommodating only a narrow range of noise levels. To counteract these biases, we have devised the following data augmentation strategies:
\begin{itemize}[noitemsep, nolistsep]
\item \textbf{Distractor Sampling}: By randomly selecting a varying number of irrelevant documents, we modify the positioning of relevant documents and the total document count within the context. This approach also mimics the fluctuating noise levels encountered in real-world scenarios, training language models to cope with noisy contexts effectively.
\item \textbf{Document Shuffling}: Reordering the documents helps to remove any superficial positional cues that could lead to reasoning bias. For example,  this ensures that models do not learn to associate the sequence of relevant documents with a fixed reasoning chain sequence.
\end{itemize}
These data augmentation strategies are applied in sequence for each training instance.

\section{Experiments}
\label{sec:exp}

\begin{table*}[t!]
    \centering
    \setlength{\tabcolsep}{3pt}
    \sisetup{detect-all}
    \resizebox{\textwidth}{!}{
    \begin{tabular}{c|l|rrrr|rS[table-format=3.0]S[table-format=3.0]S[table-format=3.0]S[table-format=3.0]|rrS[table-format=3.0]}
        \toprule
        \multicolumn{2}{c|}{\multirow{3}*{\textbf{Model}}} & \multicolumn{4}{c|}{\textbf{MuSiQue}} & \multicolumn{5}{c|}{\textbf{2Wiki}} & \multicolumn{3}{c}{\textbf{HotpotQA}} \\
        \cmidrule{3-14}
        \multicolumn{2}{c|}{} & \textbf{Overall} & \textbf{2-Hop} & \textbf{3-Hop} & \textbf{4-Hop} & \textbf{Overall} & \textbf{Compositional} & \textbf{Inference} & \textbf{Comparison} & \textbf{Bridge-Comparison} & \textbf{Overall} & \textbf{Bridge} & \textbf{Comparison} \\
        \midrule
        \multirow{20}*{\rotatebox{90}{5-Shot}} & ChatGPT & \\
        & \quad + AO & 15.8 & 16.1 & 14.3 & 17.4 & 46.2 & 19.0 & 68.1 & 30.5 & 71.4 & 51.0 & 49.0 & 60.2   \\
        & \quad + CoT & 36.2 & 34.6 & 38.3 & 37.0 & 55.2 & 24.1 & 89.1 & 28.4 & 90.5 & 56.8 & 56.1 & 60.2  \\
        & \quad + CoC & \underline{37.0} & 37.0 & 37.0 & 37.0 & \underline{55.4} & 27.8 & 89.1 & 28.4 & 88.6 & \underline{58.6} & 57.5 & 63.6 \\
        & Claude-instant & \\
        & \quad + AO & 26.2 & 25.2 & 27.3 & 27.2 & 47.0 & 19.0 & 68.9 & 32.5 & 70.5 & 54.4 & 52.7 & 62.5  \\
        & \quad + CoT & 26.0 & 27.2 & 27.9 & 19.6 & 40.8 & 20.3 & 80.7 & 15.7 & 58.1 & 20.2 & 23.5 & 4.5  \\
        & \quad + CoC & 32.2 & 32.7 & 30.5 & 33.7 & 53.4 & 36.7 & 90.8 & 22.3 & 81.9 & 54.2 & 55.3 & 48.9  \\

        & LongChat & \\
        & \quad + AO &  6.7 &  7.0 &  4.1 &  10.1 &  26.8 &  12.0 &  3.4 &  48.7 &  47.3 &  32.3 &  34.0 &  24.6  \\
        & \quad + CoT &  9.7 & 12.1 & 6.7 & 8.3 & 27.1 & 17.8 & 7.2 & 43.7 & 40.6 & 38.5 & 38.2 & 40.2 \\
        & \quad + CoC &  11.0 & 13.3 & 8.7 & 8.7 & 24.5 & 19.0 & 5.9 & 42.9 & 27.9 & 39.1 & 38.9 & 39.8 \\

        & LongLoRA & \\
        & \quad + AO & 0.2 & 0.4 & 0.0 & 0.0 & 7.7 & 9.0 & 4.6 & 13.2 & 1.3 & 16.9 & 16.3 & 19.3 \\
        & \quad + CoT & 0.0 & 0.0 & 0.0 & 0.0 & 15.1 & 5.9 & 0.8 & 28.9 & 27.3 & 11.4 & 11.0 & 13.3 \\
        & \quad + CoC & 0.0 & 0.0 & 0.0 & 0.0 & 8.3 & 3.2 & 1.3 & 15.7 & 14.6 & 4.4 & 4.0 & 6.4 \\

        & Vicuna & \\
        & \quad + AO & 0.1 & 0.1 & 0.0 & 0.0 & 20.5 & 5.2 & 5.1 & 31.9 & 47.6 & 22.3 & 24.2 & 13.6  \\
        & \quad + CoT & 0.0 & 0.0 & 0.0 & 0.0 & 27.7 & 14.7 & 7.6 & 48.5 & 43.5 & 30.5 & 32.0 & 23.1 \\
        & \quad + CoC & 0.0 & 0.0 & 0.0 & 0.0 & 28.4 & 20.6 & 7.2 & 49.3 & 35.2 & 33.1 & 34.7 & 25.8 \\

        \midrule
        \multirow{4}*{\rotatebox{90}{0-Shot}} & \method{} & \\
        & \quad + AO & 32.9 & 35.7 & 27.1 & 34.8 & \textbf{49.2} & 48.7 & 28.7 & 54.9 & 59.0 & 51.5 & 51.0 & 54.2 \\
        & \quad + CoT & 37.9 & 41.6 & 35.5 & 31.9 & 46.8 & 48.4 & 25.3 & 53.5 & 52.4 & 50.9 & 51.2 & 49.2 \\
        & \quad + CoC & \textbf{38.1} & 42.7  & 35.7  & 29.7 & 47.4  & 46.0  & 27.4  & 57.7  & 53.3  & \textbf{52.1}  & 52.6  & 49.6  \\
        \bottomrule
    \end{tabular}
    }
    \caption{Exact-Match (EM) results on three multi-hop reasoning datasets. The best small-scale long-context LM results are in \textbf{bold} and the best baseline results are \ul{underlined}.}
    \label{tab:main}
\end{table*}

\subsection{Datasets}


Our method's effectiveness in multi-hop reasoning is assessed on the following datasets: \textbf{HotpotQA} \cite{yang-etal-2018-hotpotqa}, \textbf{2WikiMultiHopQA} (\textbf{2Wiki} for short) \cite{ho-etal-2020-constructing}, and \textbf{MuSiQue} \cite{trivedi-etal-2022-musique}.
For each question, we provide a context composed of shuffled relevant and irrelevant documents. These irrelevant documents are the official retrieved distractor documents. We adopt the development and test sets from \citet{trivedi-etal-2023-interleaving} for evaluation, which contains 100 and 500 examples respectively. The results we present are the mean values from three separate trials, each with a distinct random seed.

To understand the broader impact of enhancing multi-hop reasoning on LMs' overall capabilities, we also conduct evaluations on general instruction-following benchmarks, namely \textbf{MT-Bench} \cite{DBLP:journals/corr/abs-2306-05685} and \textbf{AlpacaEval} \cite{alpaca_eval}.

\subsection{Models}

The following long-context baselines are chosen in our experiments.
\begin{itemize}[noitemsep, nolistsep]
\item \textbf{ChatGPT~\cite{DBLP:conf/nips/BrownMRSKDNSSAA20}.} We choose \texttt{gpt-3.5-turbo-1106}, which supports a window size of 16K tokens.
\item \textbf{Claude-instant~\cite{DBLP:journals/corr/abs-2204-05862}.} We choose \texttt{claude-instant-1.2}, which has a window size of 100K tokens.
\item \textbf{LongChat~\cite{longchat2023}.} We use \texttt{longchat-7b-16k}, a 7B fine-tuned LLaMA model~\cite{touvron2023llama}. It has a window size of 16K tokens.
\item \textbf{LongLoRA~\cite{chen2023longlora}.} We use \texttt{LongAlpaca-7B-16k}, which has a window size of 16K.
\item \textbf{Vicuna~\cite{DBLP:journals/corr/abs-2306-05685}.} We use \texttt{vicuna-7b-v1.5-16k}, a 7B fine-tuned LLaMA-2 model~\cite{touvron2023llama2}. It supports a window size of 16K tokens.
\end{itemize}

We prompt all models with 5-shot to evaluate their multi-hop reasoning performance.
These 5 demonstrations are randomly sampled from the 20 annotated training examples provided by \citet{trivedi-etal-2023-interleaving}.
If the input length exceeds the window size, we drop the last demonstration until the input length fits.
The prompts we used are presented in Appendix~\ref{sec:prompt}.

Our model \textbf{\method{}} is fine-tuned on \texttt{vicuna-7b-v1.5-16k} with LoRA~\cite{DBLP:conf/iclr/HuSWALWWC22}, following hyper-parameters used in FastChat~\cite{DBLP:journals/corr/abs-2306-05685}.
For its training data, we perform augmentations to double the training data for all tasks in Section~\ref{sec:learning} except for the QI task.
Note that we subsample same-sized instruction-tuning data from the Alpaca dataset~\cite{alpaca} and mix it with the reasoning data.
These instruction-tuning data serve the purpose of minimizing the risk of hampering other abilities Vicuna already possesses before fine-tuning.

\subsection{Main Results}
\label{sec:result}

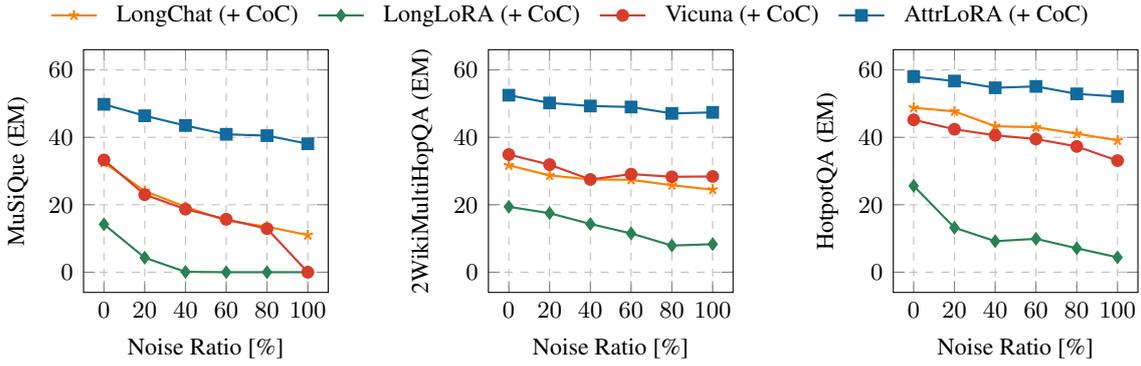
\begin{figure*}[t!]
    \centering
    \makeatletter
    \let\ref\@refstar
    \ref{grouplegend}
    \makeatother
    \begin{tikzpicture}
        \begin{groupplot}[
            group style={group size=3 by 1, horizontal sep=60pt},
            width=1.0\textwidth,
            height=0.3\textwidth,
            legend cell align={left},
            legend pos=north west,
            enlargelimits=0.1,
            legend style={
                font=\small,
                draw=none,
                column sep=5pt,
                legend columns=4,
            },
        ]
        \nextgroupplot[
            width=0.3\textwidth,height=0.3\textwidth,
            yticklabel style={/pgf/number format/fixed,/pgf/number format/precision=1},
            ylabel={MuSiQue (EM)},
            ylabel near ticks,
            xlabel={Noise Ratio [\%]},
            xlabel near ticks,
            xmajorgrids=true,
            ymajorgrids=true,
            legend pos=south west,
            grid style=dashed,
            xtick=data,
            every tick label/.append style={font=\small},
            label style={font=\small},
            ylabel style={yshift=0pt},
            legend to name=grouplegend,
            xmin=0,xmax=100,
            ymax=60,ymin=0,
        ]
            \addplot [orange,thick,mark=star] coordinates {
                (0,32.5) (20,24.1) (40,19.4) (60, 15.3)(80, 13.5)(100, 11.0)
            };\addlegendentry{LongChat (+ CoC)}
            \addplot [lyygreen,thick,mark=diamond*] coordinates {
                (0,14.2) (20,4.3) (40,0.1) (60,0.0) (80,0.0) (100,0.0)
            };\addlegendentry{LongLoRA (+ CoC)}
            \addplot [lyyred,thick,mark=*] coordinates {
                (0,33.3) (20,23.0) (40,18.7) (60,15.7) (80,12.9) (100,0.0)
            };\addlegendentry{Vicuna (+ CoC)}
            \addplot [lyyblue,thick,mark=square*] coordinates {
                (0,49.8) (20,46.4) (40,43.5) (60,40.9) (80,40.5) (100,38.1)
            };\addlegendentry{\method{} (+ CoC)}
            
        \nextgroupplot[
            width=0.3\textwidth,height=0.3\textwidth,
            yticklabel style={/pgf/number format/fixed,/pgf/number format/precision=1},
            ylabel={2WikiMultiHopQA (EM)},
            ylabel near ticks,
            xlabel={Noise Ratio [\%]},
            xlabel near ticks,
            xmajorgrids=true,
            ymajorgrids=true,
            grid style=dashed,
            xtick=data,
            every tick label/.append style={font=\small},
            label style={font=\small},
            ylabel style={yshift=0pt},
            xmin=0,xmax=100,
            ymax=60,ymin=0,
        ]
            \addplot [orange,thick,mark=star] coordinates {
                (0,31.7) (20,28.7) (40,27.5) (60, 27.4) (80, 25.8)(100, 24.5)
            };
            \addplot [lyygreen,thick,mark=diamond*] coordinates {
                (0,19.4) (20,17.5) (40,14.3) (60, 11.5) (80, 7.9) (100, 8.3)
            };
            \addplot [lyyred,thick,mark=*] coordinates {
                (0,34.9) (20,31.9) (40,27.5) (60, 29.1)(80, 28.3)(100, 28.4)
                
            };
            \addplot [lyyblue,thick,mark=square*] coordinates {
                (0,52.5) (20,50.2) (40,49.3) (60, 49.0)(80, 47.1)(100, 47.4)
                
            };
            
        \nextgroupplot[
            width=0.3\textwidth,height=0.3\textwidth,
            yticklabel style={/pgf/number format/fixed,/pgf/number format/precision=1},
            ylabel={HotpotQA (EM)},
            ylabel near ticks,
            xlabel={Noise Ratio [\%]},
            xlabel near ticks,
            xmajorgrids=true,
            ymajorgrids=true,
            grid style=dashed,
            xtick=data,
            every tick label/.append style={font=\small},
            label style={font=\small},
            ylabel style={yshift=0pt},
            xmin=0,xmax=100,
            ymax=60,ymin=0,
        ]
            \addplot [orange,thick,mark=star] coordinates {
                (0,48.8) (20,47.7) (40,43.3) (60, 43.0) (80, 41.1) (100, 39.1)
            };
            \addplot [lyygreen,thick,mark=diamond*] coordinates {
                (0,25.6) (20,13.2) (40,9.2) (60, 9.9) (80, 7.1) (100, 4.4)
            };
            \addplot [lyyred,thick,mark=*] coordinates {
                (0,45.2) (20,42.4) (40,40.6) (60, 39.5) (80, 37.3) (100, 33.1)
                
            };
            \addplot [lyyblue,thick,mark=square*] coordinates {
                (0,58.0) (20,56.7)  (40,54.7) (60,55.1) (80,52.9) (100,52.1)
            };
            
        \end{groupplot}
    \end{tikzpicture}
    \caption{Exact-Match (EM) results of different models under various noise levels in three multi-hop reasoning datasets. Note that all models except our \method{} use 5-shot prompting. A higher noise ratio indicates more distractors, i.e., irrelevant documents, are presented in the context of both the test instance and the demonstrations.}
    \label{fig:noise}
\end{figure*}



\noindent\textbf{Effectiveness and Robustness of Reasoning with Attributions.}
The results in Table \ref{tab:main} underscore the efficacy of our CoC prompting across three multi-hop reasoning datasets, benchmarked against five baselines. In 77\% of the evaluated cases (disregarding instances of near-zero model performance) CoC outperforms CoT. Notably, Claude-instant exhibits strong results with AO, and its performance diminishes when CoT is used. However, CoC not only mitigates this decline but also attains results on par with AO, demonstrating the robustness of attribution-based reasoning. For a detailed analysis of CoC's robustness, particularly against varying degrees of contextual noise, see Appendix \ref{sec:robust}.




\noindent\textbf{Performance of \method{} Against Proprietary Models.}
Table \ref{tab:main} presents a zero-shot performance comparison between our \method{} and five-shot outcomes from various baselines. \method{} surpasses baselines of comparable scale by an average margin of over 20 points. It exceeds the performance of two notable proprietary models on MuSiQue and delivers closely competitive results on the other two benchmarks.

\begin{table}[t!]
    \centering
    {\small
    \begin{tabular}{l|rr}
        \toprule
        \makecell[c]{\textbf{Model}} & \makecell[c]{\textbf{MT-Bench}\\(Score)} & \makecell[c]{\textbf{AlpacaEval}\\(Win Rate)} \\
        \midrule
        ChatGPT & 8.245 & 9.178\% \\
        Claude-instant & 8.131 & 15.664\% \\
        Vicuna & 6.068 & 5.415\% \\
        \quad + Alpaca Data & 4.850 & 3.287\% \\
        \method{} & 4.978 & 3.106\% \\
        \bottomrule
    \end{tabular}
    }
    \caption{Results on general instruction-following benchmarks. ``+ Alpaca Data'' is a Vicuna-7B model continued fine-tuning on Alpaca data.}
    \label{tab:general}
\end{table}

In particular, \method{} with AO achieves superior results on 2Wiki and surpasses CoT on HotpotQA. This can be attributed to the relative simplicity of these datasets, where explicit reasoning does not significantly enhance performance. For instance, CoT's advantage is noticeably smaller on these datasets compared to MuSiQue for both the ChatGPT and Claude-instant. Additionally, according to \citet{jiang-bansal-2019-avoiding}, over half of the ``bridge-type'' questions in HotpotQA contain shortcuts, which can locate the answer by keyword matching, circumventing the need for the intended two-hop reasoning. Similarly, 2Wiki's predictable nature, due to its question construction from a limited set of rules, simplifies the task for LMs.
Another contributing factor is that \method{} is trained on \dataset{}, which does not encompass the full range of question types found in 2Wiki and HotpotQA, such as ``comparison-type'' questions.



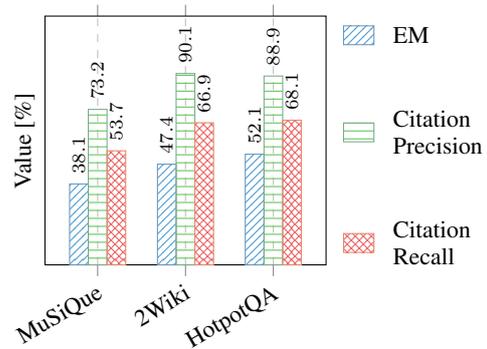
\begin{figure}[t!]
    \centering
    \begin{tikzpicture}
        \begin{axis}[
            width=0.33\textwidth,height=0.305\textwidth,
            legend cell align={left},
            legend pos=north west,
            enlargelimits=0.1,
            legend pos=outer north east,
            legend style={
                font=\small,
                draw=none,
                column sep=5pt,
                /tikz/every even column/.append style={column sep=30pt},
                legend columns=1,
                row sep=20pt,
                cells={align=left},
            },
            symbolic x coords={MuSiQue,2Wiki,HotpotQA},
            enlarge x limits=0.3,
            enlarge y limits={upper,value=0.3},
            ylabel={Value [\%]},
            ylabel near ticks,
            ybar=0pt,
            xtick=data,
            ytick=\empty,
            ymin=0,
            nodes near coords,
            every node near coord/.append style={rotate=90,anchor=west,font=\scriptsize},
            every tick label/.append style={font=\footnotesize},
            xticklabel style={rotate=30,anchor=north east,font=\footnotesize,inner sep=0pt,outer sep=2pt},
            bar width=7pt,
            xmajorgrids=true,
            ymajorgrids=true,
            grid style=dashed,
            label style={font=\small},
            legend image code/.code={\draw [#1] (0cm,-0.1cm) rectangle ++(0.4cm,0.25cm);},
        ]
            \addplot [draw=lyyblue!60,fill=lyyblue!60,pattern=north east lines,pattern color=lyyblue!60] coordinates {(MuSiQue,38.1) (2Wiki,47.4) (HotpotQA,52.1)};
            \addlegendentry{EM}
            \addplot [draw=lyygreen!60,fill=lyygreen!60,pattern=horizontal lines,pattern color=lyygreen!60] coordinates {(MuSiQue,73.2) (2Wiki,90.1) (HotpotQA,88.9)};
            \addlegendentry{Citation\\Precision}
            \addplot [draw=lyyred!60,fill=lyyred!60,pattern=crosshatch,pattern color=lyyred!60] coordinates {(MuSiQue,53.7) (2Wiki,66.9) (HotpotQA,68.1)};
            \addlegendentry{Citation\\Recall}
        \end{axis}
    \end{tikzpicture}
    \caption{Multi-hop reasoning performance vs. citation precision and recall of \method{}.}
    \label{fig:citation}
\end{figure}

\noindent\textbf{Resilience of \method{} to Noisy Contexts.}
A key aspect of \method{} is its robustness to contextual noise. To investigate this, Figure \ref{fig:noise} illustrates \method{}'s performance against varying degrees of synthesized noise. This synthesized noise is implemented by adding varied numbers of random irrelevant documents to the context. The data indicates that while the performance of baseline models markedly declines with increased noise, e.g., Vicuna drops by over 30 points on MuSiQue, \method{} shows greater resilience, with a reduction of only about 10 points.



\noindent\textbf{Impact of Attribution Learning on General Abilities.}
Our investigation extends beyond multi-hop reasoning to examine how attribution learning affects \method{}'s general instruction-following capabilities post-fine-tuning, as compared to the Vicuna baseline. The results in Table \ref{tab:general} from two instruction-following benchmarks reveal that fine-tuning slightly compromises abilities beyond multi-hop reasoning in a 7B model due to capacity constraints. However, a closer analysis reveals that over 98\% of the performance decrease is attributed to fine-tuning with Alpaca data (``+ Alpaca Data''), while multi-hop reasoning data incurs less than a 2\% detriment. This is because the quality of Alpaca data is inferior to Vicuna's, with the former being single-turn GPT-3 synthesized and the latter comprising multi-turn human-bot conversations.

\subsection{Analysis}
\label{sec:analysis}



\begin{table}[t!]
    \centering
    \setlength{\tabcolsep}{3pt}
    {\small
    \begin{tabular}{l|ccc}
        \toprule
        \makecell[c]{\textbf{Model}} & \textbf{MuSiQue} & \textbf{2Wiki} & \textbf{HotpotQA} \\
        \midrule
        Vicuna (5-Shot) & 0.00 & 28.4 & 33.1  \\
        \quad + AP & 32.3 & 47.8 & 52.1 \\
        \quad + CG & 37.3 & 45.3 & 50.7 \\
        \quad + LA & 37.3 & 46.9 & 52.1 \\
        \quad + QI & 38.1 & 47.4 & 52.1 \\
        \bottomrule
    \end{tabular}
    }
    \caption{Ablation study on multi-task learning.}
    \label{tab:task}
\end{table}

\begin{table}[t!]
    \centering
    \setlength{\tabcolsep}{4pt}
    {\small
    \begin{tabular}{l|ccc}
        \toprule
        \makecell[c]{\textbf{Model}} & \textbf{MuSiQue} & \textbf{2Wiki} & \textbf{HotpotQA} \\
        \midrule
        Vicuna (5-Shot) & 0.00 & 28.4 & 33.1 \\
        \midrule
        \quad + AP & 27.4 & 44.9 & 50.9\\
        \quad + Augmentation & 32.3 & 47.8 & 52.1\\
        \midrule
        \quad + CG & 28.1 & 37.7 & 49.4 \\
        \quad + Augmentation & 30.5 & 37.3 & 50.4\\
        \midrule
        \quad + LA & 29.1 & 38.8 & 49.0 \\
        \quad + Augmentation & 30.1 & 37.1 & 49.9 \\
        \bottomrule
    \end{tabular}
    }
    \caption{Ablation study on data augmentation.}
    \label{tab:aug}
\end{table}

\noindent\textbf{Attribution Quality of \method{}.}
Drawing from the insights of \citet{gao-etal-2023-enabling}, we scrutinize the citation precision and recall for \method{}, as presented in Figure \ref{fig:citation}. The model demonstrates high precision, indicating its proficiency in correctly attributing statements to pertinent documents. Nonetheless, the moderate recall highlights that \method{} does not consistently identify all relevant documents, a potential consequence of the disconnected reasoning patterns observed in \dataset{} (detailed in Appendix \ref{sec:human}). Further exploration of the correlation between attribution quality and reasoning performance is in Appendix \ref{sec:attribute}.

\noindent\textbf{The Effectiveness of Multi-Task Learning.}
Our ablation study in Table \ref{tab:task} assesses our multi-task learning approach. Results indicate a marked enhancement in Vicuna's reasoning capabilities upon fine-tuning with our dataset (``+ AP''). However, explicitly training Vicuna to generate CoT (``+ CG'') yields mixed outcomes: It benefits performance on MuSiQue but adversely affects results on 2Wiki and HotpotQA. This discrepancy can be attributed to the relative ease of the latter datasets, where simpler questions and shortcuts reduce the effectiveness of complex reasoning strategies, as discussed in Section \ref{sec:result}.
Importantly, integrating the LA task (``+ LA'') mitigates the performance drops associated with CoT and notably boosts MuSiQue scores. This implies that attributions are instrumental in enabling the model to reason over complicated questions without compromising its ability to handle simpler queries.
Finally, the addition of the QI task (``+ QI'') appears to further refine the model's multi-hop reasoning proficiency, underscoring the value of our multi-task learning framework.

\begin{figure}[t!]
    \centering
    \makeatletter
    \let\ref\@refstar
    \ref{outsidelegend}
    \makeatother
    \\
    \begin{tikzpicture}
        \begin{axis}[
            width=0.33\textwidth,height=0.305\textwidth,
            legend cell align={left},
            legend pos=north west,
            enlargelimits=0.1,
            legend style={
                font=\small,
                draw=none,
                column sep=3pt,
                legend columns=3,
                cells={align=left},
            },
            legend to name=outsidelegend,
            yticklabel style={/pgf/number format/fixed,/pgf/number format/precision=1},
            ylabel={EM},
            ylabel near ticks,
            xlabel={Training Set Size [\%]},
            xlabel near ticks,
            xmajorgrids=true,
            ymajorgrids=true,
            grid style=dashed,
            xtick={0, 20,40,...,100},
            xmin=0,xmax=100,
            every tick label/.append style={font=\small},
            label style={font=\small},
            ylabel style={yshift=0pt},
        ]
            \addplot [lyygreen,thick,mark=triangle*] coordinates {
                (0,0)
                (20,30.2)
                (40,34.9)
                (60,35.4)
                (80,36.9)
                (100,38.1)
                
            };\addlegendentry{MuSiQue}
            \addplot [lyyred,thick,mark=square*] coordinates {
                (0,28.4)
                (20,38.9)
                (40,43.5)
                (60,46.2)
                (80,44.9)
                (100,47.4)
            };\addlegendentry{2Wiki}
            \addplot [lyyblue,thick,mark=*,mark options={solid}] coordinates {
                (0,33.1)
                (20,49.0)
                (40,51.1)
                (60,53.1)
                (80,52.2)
                (100,51.7)
            };\addlegendentry{HotpotQA}
        \end{axis}
    \end{tikzpicture}
    \caption{The impact of scaling fine-tuning data size.}
    \label{fig:size}
\end{figure}
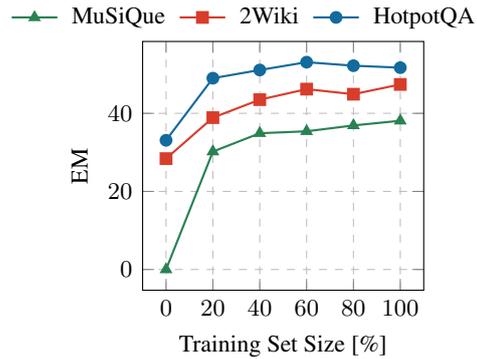

\begin{table*}[t!]
    \centering
    \small
    \setlength{\tabcolsep}{1.5mm}{\scalebox{0.97}{
    \begin{tabular}{p{7cm}|p{8.5cm}}
    \toprule
    \textbf{Question}: Who is the mascot of the university related to Randy Conrads? & 
    \textbf{Question}: What is the record label of the co-writer and recording artist of Permission to Fly?\\
    \textbf{Document} [4](Title: Benny Beaver): \blue{Benny Beaver is the official mascot of Oregon State University} and winner of the 2011 Capital One Mascot of the Year write - in campaign. \ldots & 
    \textbf{Document} [8](Title: No Ordinary Girl): \blue{No Ordinary Girl is the debut album by Jordan Pruitt, released in the United States on February 6, 2007 by Hollywood Records}. The album debuted and peaked at number sixty-four \ldots\\
    \textbf{Document} [7](Title: Randy Conrads): \blue{Randy Conrads attended Oregon State University}, graduating in 1972 with a bachelor's degree in industrial engineering. \ldots & 
    \textbf{Document} [19](Title: Permission to Fly): \blue{Permission to Fly is the second and final studio album by American singer-songwriter Jordan Pruitt}. On July 22, 2008, the album was released \ldots\\
    \gray{(Other irrelevant documents are omitted.)} & \gray{(Other irrelevant documents are omitted.)} \\
    \textbf{Vicuna}: \red{the university. \ \xmark} & 
    \textbf{Vicuna}: The answer is: \red{The answer is a question. \ \xmark}\\
    \textbf{\method{}}: Randy Conrads attended Oregon State University \org{[7]}. The mascot of Oregon State University is Benny Beaver \org{[4]}. The answer is: \gr{Benny Beaver \ \cmark} & 
    \textbf{\method{}}: The co-writer and recording artist of ``Permission to Fly'' is Jordan Pruitt \org{[19]}. The record label of Jordan Pruitt is Hollywood Records \org{[8]}. The answer is: \gr{Hollywood Records \ \cmark}\\
    \bottomrule
    \end{tabular}}}
    \caption{Two examples from MuSiQue test set where Vicuna fails and \method{} succeeds. We manually annotate quotes in \blue{blue} and citations in \org{orange}. Correct predictions are marked by \gr{\cmark} and incorrect ones are by \red{\xmark}.}
    \label{tab:case}
\end{table*}

\noindent\textbf{The Effectiveness of Data Augmentation.}
We explore the impact of our data augmentation strategy, intentionally omitting the QI task, as it alone is insufficient for training models to conduct multi-hop reasoning. The data in Table \ref{tab:aug} demonstrates that including augmented data generally enhances model performance across various datasets. However, augmenting CG and LA data does not yield improvements on 2Wiki. In this case, the model readily learns from a limited amount of annotated data due to the simplicity of the automatically generated questions within 2Wiki.
Conversely, on MuSiQue and HotpotQA, which feature more complex and varied human-crafted questions, the model benefits from exposure to a larger dataset to accommodate the diversity of question formulations.

\noindent\textbf{The Effectiveness of Scaling Fine-Tuning Data.}
In Figure \ref{fig:size}, we investigate how the expansion of fine-tuning data influences model performance. It is evident that incorporating additional data steadily enhances performance on MuSiQue and 2Wiki, while optimal results are attained with just 60\% of our data for HotpotQA. This fact suggests that more complex question answering, involving additional reasoning steps like MuSiQue and 2Wiki, demands a larger dataset.
An intriguing discovery is that using a mere 20\% of our data achieves approximately 85\% of the peak performance. This highlights the efficiency of fine-tuning: Even a modest subset of multi-hop reasoning examples can significantly boost the model's reasoning capabilities.

\noindent\textbf{Case Study.}
Table \ref{tab:case} presents a comparative case study where Vicuna and \method{} are both prompted to generate CoC. Within the provided examples, \method{} successfully produces coherent CoT and precisely attributes each claim. In contrast, Vicuna yields answers without engaging in an explicit reasoning process.

\section{Related Work}


\paragraph{Multi-Hop Reasoning.}



Multi-hop reasoning in open-domain question answering requires the synthesis and analysis of disparate facts across various documents to formulate a response. Key datasets in this field include HotpotQA \cite{yang-etal-2018-hotpotqa}, 2Wiki \cite{ho-etal-2020-constructing}, and MuSiQue \cite{trivedi-etal-2022-musique}, which predominantly adopt a reading comprehension framework with pre-retrieved documents supplied by the creators.
Traditional approaches often utilize a selector-reader model \cite{zhang2023beam,zhu2021retrieving}, where the selector is tasked with pinpointing relevant documents from the provided set, and the reader constructs an answer based on these selections.

Recent advances, however, pivot towards a paradigm that leverages long-context LMs \cite{DBLP:conf/iclr/KhotTFF0CS23,trivedi-etal-2023-interleaving}. In this approach, the role of the selector is phased out, and instead, the entirety of the retrieved documents is processed by a long-context LM, which acts as the reader.
Our study aligns with this emergent research trend, particularly focusing on the use of attributions to enhance the performance of multi-hop reasoning within this long-context LM framework.

\paragraph{Context Utilization.}


The recent advent of long-context LMs has shown promise \cite{longchat2023,DBLP:journals/corr/abs-2306-05685,chen2023longlora}. However, these models often struggle with noisy contexts. \citet{DBLP:conf/icml/ShiCMSDCSZ23} demonstrate that superfluous sentences can significantly disrupt mathematical reasoning. \citet{liu2023lost} identify a relevant document position bias in multi-document QA. \citet{wu2024easily} show that LMs could be easily distracted by retrieved irrelevant inputs.

To mitigate the impact of irrelevant context, \citet{DBLP:conf/icml/ShiCMSDCSZ23} prompt models to disregard such information and adopt self-consistency techniques \cite{DBLP:conf/iclr/0002WSLCNCZ23}. \citet{DBLP:conf/iclr/CreswellSH23} suggest a two-stage approach that focuses on fact selection prior to reasoning. Echoing this approach, \citet{DBLP:journals/corr/abs-2311-09210} introduce Chain-of-Note which entails reviewing document relevance before providing an answer. Meanwhile, \citet{DBLP:journals/corr/abs-2310-01558} examine automatic data generation for training more robust models.
Our research contributes to this domain by investigating the use of attributions as a novel method for effective context utilization.

\paragraph{Language Models Attribution.}


Attribution in language models constitutes a nascent area of study, primarily aimed at identifying and mitigating hallucination \cite{li2023survey}. A line of research concentrates on post-retrieval answering: Models provide responses based on retrieved results with cited attributions \cite{DBLP:journals/corr/abs-2112-09332, DBLP:journals/corr/abs-2203-11147, gao-etal-2023-enabling}.
Our research emerges from this foundation but diverges in its application; We focus on multi-hop reasoning rather than hallucination reduction. Moreover, we delve into optimizing training methodologies to maximize the efficacy of scarce attribution annotations.

\section{Conclusion}


This study demonstrates that long-context LMs face challenges with multi-hop reasoning within noisy contexts. We introduce a reasoning paradigm that incorporates attributions, which significantly improves the reasoning capabilities of long-context LMs. Alongside, we contribute a new dataset annotated with attributions and study training strategies tailored for multi-hop reasoning. Our comprehensive experiments across five models and five benchmarks validate the superiority of our approach in enhancing multi-hop reasoning performance.

\section*{Limitations}


The proposed reasoning with attributions method currently leverages citations and quotations, leaving room for future exploration of other attribution forms, such as URLs. This work concentrates on refining training strategies, yet the potential of custom model architectures for this task remains untapped. Additionally, our approach relies on contexts pre-supplied by dataset creators. Emerging research suggests that language models integrated with search engines can achieve enhanced outcomes. A promising avenue for further research lies in developing models that not only manage noisy contexts more effectively, as demonstrated in our work, but also actively engage with search tools to improve information retrieval.

\section*{Ethics Statement}

Our dataset builds upon MuSiQue, adhering to its original copyright provisions; We distribute our supplementary annotations under the CC BY 4.0 license. While our model-generated annotations could potentially include misinformation or harmful content, our manual quality review in Appendix~\ref{sec:human} did not encounter such instances. Throughout the human annotation phase, we did not gather any demographic data or information that could reveal the identity of the annotators. All annotators provided informed consent for the use of their annotations exclusively for research purposes.

\section*{Acknowledgements}
This work was supported by National Key R\&D Program of China (Project No. 2022ZD0161200, 2022ZD0161201).
This work is also supported by Hong Kong Research Grant Council - Early Career Scheme (Grant No. 24200223) and Hong Kong Innovation and Technology Commission Project No. ITS/228/22FP.
This work was also partially funded by the Centre for Perceptual and Interactive Intelligence (CPII) Ltd under the Innovation and Technology Commission (ITC)'s InnoHK. 

\bibliography{custom}

\clearpage

\appendix

\begin{figure*}
    \centering
    \includegraphics[width=0.8\linewidth]{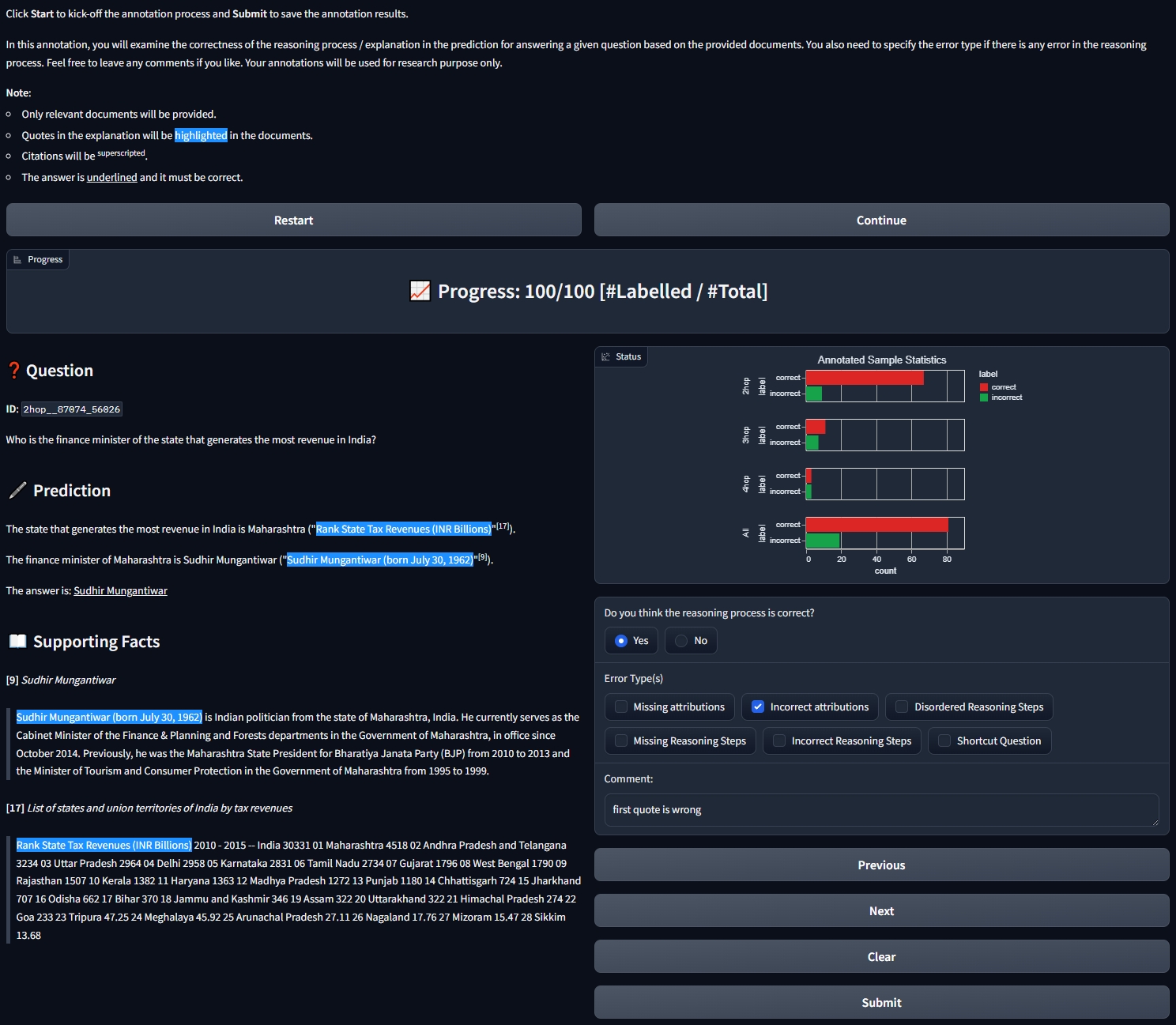}
    \caption{Screenshot of our human annotation tool.}
    \label{fig:tool}
\end{figure*}

\section{Human Assessment of \dataset{}}
\label{sec:human}



We recruit a student possessing expertise in NLP and holding a Master's degree in Computer Science to annotate 100 random samples from our \dataset{}.
The tool we used for annotations is shown in Figure~\ref{fig:tool}.
The annotation results reveal that 19\% of the CoT annotations exhibited reasoning issues despite correct final answers, a problem referred to as \textit{reasoning unfaithfulness} \cite{DBLP:journals/corr/abs-2208-14271}. 
These erroneous CoTs fall into three categories: 5.26\% with \texttt{Disordered Steps} where reasoning steps are improperly arranged, 78.95\% with \texttt{Missing Steps} indicating omitted essential reasoning steps, also known as \textit{disconnected reasoning} \cite{trivedi-etal-2020-multihop}, and 10.53\% with \texttt{Incorrect Steps} containing invalid reasoning steps. Our assessment also identified that 9\% of questions in the dataset allowed for \textit{shortcuts} \cite{jiang-bansal-2019-avoiding}, enabling models to find answers by keyword matching without reasoning.

The study further showed that within incorrect CoT annotations, 47.37\% pertained to 2-hop questions, 36.84\% to 3-hop, and 15.79\% to 4-hop questions. Considering the distribution of hops in our dataset, the unfaithfulness issue affected 11.84\% of 2-hop CoTs, 38.89\% of 3-hop CoTs, and 50\% of 4-hop CoTs, suggesting a decrease in reliability of LMs with the increase in reasoning steps required.

\section{Attribution Quality and Reasoning Performance}
\label{sec:attribute}

\input{figures/cite_vs_em}

\begin{table}[t!]
    \centering
    \setlength{\tabcolsep}{4pt}
    {\small
    \begin{tabular}{l|rrr}
        \toprule
        \makecell[c]{\textbf{Entry}} & \makecell[c]{\textbf{Pearson}} & \makecell[c]{\textbf{Spearman}} & \makecell[c]{\textbf{Kendall}} \\
        \midrule
        \multicolumn{4}{c}{\cellcolor{lightgray}\textit{MuSiQue}} \\
        \midrule
        EM vs. P & 0.887\rlap{$^*$} & 0.951\rlap{$^*$} & 0.826\rlap{$^*$} \\
        EM vs. R & 0.484\rlap{$^*$} & 0.481\rlap{$^*$} & 0.411\rlap{$^*$} \\
        \midrule
        \multicolumn{4}{c}{\cellcolor{lightgray}\textit{2Wiki}} \\
        \midrule
        EM vs. P & 0.917\rlap{$^*$} & 0.896\rlap{$^*$} & 0.766\rlap{$^*$} \\
        EM vs. R & -0.172 & -0.052 & 0.024 \\
        \midrule
        \multicolumn{4}{c}{\cellcolor{lightgray}\textit{HotpotQA}} \\
        \midrule
        EM vs. P & 0.865\rlap{$^*$} & 0.868\rlap{$^*$} & 0.741\rlap{$^*$} \\
        EM vs. R & -0.275 & -0.038 & 0.023 \\
        \bottomrule
    \end{tabular}
    }
    \caption{Correlation coefficients between citation precision (denoted as \textbf{P}) or recall (denoted as \textbf{R}) and the multi-hop reasoning performance (\textbf{EM}). $^*$ denotes statistically significant ($p<0.05$).}
    \label{tab:correlation}
\end{table}


Figure~\ref{fig:cite_vs_em} presents a visualization of the relationship between citation precision or recall and multi-hop reasoning performance. Our analysis identifies a notable positive correlation between citation precision and multi-hop reasoning capabilities. Table~\ref{tab:correlation} provides statistical support for this observation, with correlation coefficients indicating a significant positive correlation. These results suggest the potential of using citation precision as a reference-free proxy for assessing multi-hop reasoning performance, which could be facilitated by Natural Language Inference (NLI) methods \cite{gao-etal-2023-enabling}.

\section{Robustness to Noisy Context}
\label{sec:robust}


We assessed the resilience of CoT and CoC against varying degrees of contextual noise by evaluating the performance variability of different models across multi-hop reasoning benchmarks. Table~\ref{tab:range} indicates that in 84\% of the cases, CoC exhibits a smaller performance range compared to CoT. Notably, even in instances where CoC demonstrates a greater range, the disparity with CoT remains marginal. These findings suggest that CoC is generally more robust to noisy contexts than CoT.

\begin{table}[t!]
    \centering
    \setlength{\tabcolsep}{4pt}
    {\small
    \begin{tabular}{l|rr|rr|rr}
        \toprule
        \makecell[c]{\multirow{3}*{\textbf{Model}}} & \multicolumn{2}{c|}{\textbf{MuSiQue}} & \multicolumn{2}{c|}{\textbf{2Wiki}} & \multicolumn{2}{c}{\textbf{HotpotQA}} \\
        \cmidrule{2-7}
        & \textbf{CoT} & \textbf{CoC} & \textbf{CoT} & \textbf{CoC} & \textbf{CoT} & \textbf{CoC} \\
        \midrule
        LongChat & 22.3 & \textbf{21.5} & \textbf{6.9} & 7.2 & 9.7 & \textbf{9.7} \\
        LongLoRA & 24.7 & \textbf{14.2} & 16.4 & \textbf{11.5} & 33.5 & \textbf{21.2} \\
        Vicuna & \textbf{32.9} & 33.3 & 10.1 & \textbf{7.5} & 16.3 & \textbf{12.1} \\
        \method{} & 12.1 & \textbf{11.7} & 7.9 & \textbf{5.4} & 6.0 & \textbf{5.9} \\
        \bottomrule
    \end{tabular}
    }
    \caption{The performance range of different models in three multi-hop reasoning benchmarks when the noise ratio of the context goes from 0\% to 100\% and models are prompted with CoT or CoC. The more robust results are highlighted in \textbf{bold}.}
    \label{tab:range}
\end{table}

\section{Prompting Details}
\label{sec:prompt}

In this section, we highlight some details when prompting long-context LMs with various strategies, e.g., AO, CoT, CoC and CoQ.
\begin{itemize}[noitemsep, nolistsep]
\item Each prompting strategy has its own instructions.
\item For CoT, CoC and CoQ, we add ``Think step-by-step.'' to the end of the question.
\item For few-shot prompting, we put demonstrations to different turns as the dialogue history.
\item Each demonstration is formatted according to Table~\ref{tab:example}.
\end{itemize}

The instruction for AO is (unique descriptions are \ul{underlined}):
\begin{quote}
    \textit{Write an accurate and concise answer for the given question using only the provided search results (some of which might be irrelevant). \ul{Do not say anything other than the answer itself.}}
\end{quote}

The instruction for CoT is (differences to AO are \ul{underlined}):
\begin{quote}
    \textit{Write an accurate and concise answer for the given question using only the provided search results (some of which might be irrelevant). \ul{Start with an accurate, engaging, and concise explanation based only on the provided documents. Must end with ``The answer is:''. Use an unbiased and journalistic tone.}}
\end{quote}

The instruction for CoC is (differences to CoT are \ul{underlined}):
\begin{quote}
    \textit{Write an accurate and concise answer for the given question using only the provided search results (some of which might be irrelevant) \ul{and cite them properly}. Start with an accurate, engaging, and concise explanation based only on the provided documents. Must end with ``The answer is:''. Use an unbiased and journalistic tone. \ul{Always cite for any factual claim.}}
\end{quote}

The instruction for CoQ is (differences to CoC are \ul{underlined}):
\begin{quote}
    \textit{Write an accurate and concise answer for the given question using only the provided search results (some of which might be irrelevant) and cite them properly. Start with an accurate, engaging, and concise explanation based only on the provided documents. Must end with ``The answer is:''. Use an unbiased and journalistic tone. Always cite \ul{and extract word-for-word quotes} for any factual claim.}
\end{quote}

\section{Filtering Implementation}
\label{sec:implement}

The implementation of each filtering strategy is detailed as follows:
\begin{itemize}
    \item \textbf{Incorrect Answer}: Leveraging the evaluation script from QuAC\footnote{\url{https://s3.amazonaws.com/my89public/quac/scorer.py}}, we normalize both model predictions and official reference answers. We filter out data instances where the model prediction does not exactly match the answer.
    \item \textbf{Non-Existent Attributions}: Utilizing regular expressions, we extract all citations and verify if the predicted citations are present in the context. Each quote is checked for its exact presence in the cited document, and instances with fabricated citations or quotes are removed.
    \item \textbf{Incorrect Citations}: We assess the correctness of predicted citations using officially annotated supporting documents, ensuring the cited document is among the supporting documents. Any example with at least one incorrect citation is deleted.
    \item \textbf{Repeated Citations}: Regular expressions are used to extract all citations, and we check for duplicates. Examples with any duplicate citations are discarded.
    \item \textbf{Extreme Quotes}: We extract all quotes using regular expressions and tokenize them with NLTK\footnote{\url{https://www.nltk.org/}}. Examples are kept only if all quotes contain more than five words but do not span entire cited documents.
\end{itemize}

\section{Training and Inference Cost}


For fine-tuning using LoRA \cite{DBLP:conf/iclr/HuSWALWWC22}, we employed ZeRO-3 \cite{DBLP:conf/sc/RajbhandariRRSH21} optimization and gradient checkpointing \cite{DBLP:journals/corr/ChenXZG16} techniques. The model was trained on 8 NVIDIA A100 80GB GPUs for no more than 14 hours. Approximately 4M parameters were tuned, constituting 6.22\% of the total parameters. For inference, each test set was processed in about 30 minutes using a single NVIDIA A100 80GB GPU.

\end{document}